\documentclass{article}

\usepackage{microtype}
\usepackage{graphicx}
\usepackage{booktabs} 
\usepackage{url}            
\usepackage{booktabs}       
\usepackage{amsfonts}       
\usepackage{nicefrac}       
\usepackage{microtype}      
\usepackage[english]{babel}
\usepackage{graphicx}
\usepackage{framed}
\usepackage[normalem]{ulem}
\usepackage{bm}
\usepackage{amsmath}
\usepackage{amsthm}
\usepackage{amssymb}
\usepackage{amsfonts}
\usepackage{enumerate}
\usepackage{dsfont}
\usepackage{xcolor}
\usepackage{subfig}
\usepackage{comment}
\usepackage{tikz}
\usepackage{bibunits}
\defaultbibliography{bibliography}

\usetikzlibrary{cd,arrows,calc,shapes,positioning}
\tikzstyle{obs} = [circle,fill=white,draw=black,inner sep=1pt,minimum size=20pt,font=\fontsize{10}{10}\selectfont,node distance=1,thick]
\tikzstyle{latent} = [obs,dotted]
\tikzstyle{plate} = [draw, rectangle, minimum size=20pt]

\definecolor{CGreen}{HTML}{469B4B}
\definecolor{CBlue}{HTML}{3772FF}
\definecolor{CRose}{HTML}{DF2935}
\definecolor{CYellow}{HTML}{FDCA40}
\definecolor{CViolet}{HTML}{9000A3}
\definecolor{CCoral}{HTML}{F78154}

\newcommand{\edge}[3][]{ %
  \foreach \x in {#2} { %
    \foreach \y in {#3} { %
      \path (\x) edge [->, >={triangle 45}, #1,thick] (\y) ;%
    } ;
  } ;
}
\newcommand{\E}{\ensuremath \mathbb{E}}

\makeatletter
\newcommand*{\da@rightarrow}{\mathchar"0\hexnumber@\symAMSa 4B }
\newcommand*{\da@leftarrow}{\mathchar"0\hexnumber@\symAMSa 4C }
\newcommand*{\xdashrightarrow}[2][]{%
  \mathrel{%
    \mathpalette{\da@xarrow{#1}{#2}{}\da@rightarrow{\,}{}}{}%
  }%
}
\newcommand{\xdashleftarrow}[2][]{%
  \mathrel{%
    \mathpalette{\da@xarrow{#1}{#2}\da@leftarrow{}{}{\,}}{}%
  }%
}
\newcommand*{\da@xarrow}[7]{%
  \sbox0{$\ifx#7\scriptstyle\scriptscriptstyle\else\scriptstyle\fi#5#1#6\m@th$}%
  \sbox2{$\ifx#7\scriptstyle\scriptscriptstyle\else\scriptstyle\fi#5#2#6\m@th$}%
  \sbox4{$#7\dabar@\m@th$}%
  \dimen@=\wd0 %
  \ifdim\wd2 >\dimen@
    \dimen@=\wd2 %
  \fi
  \count@=2 %
  \def\da@bars{\dabar@\dabar@}%
  \@whiledim\count@\wd4<\dimen@\do{%
    \advance\count@\@ne
    \expandafter\def\expandafter\da@bars\expandafter{%
      \da@bars
      \dabar@ 
    }%
  }%
  \mathrel{#3}%
  \mathrel{%
    \mathop{\da@bars}\limits
    \ifx\\#1\\%
    \else
      _{\copy0}%
    \fi
    \ifx\\#2\\%
    \else
      ^{\copy2}%
    \fi
  }%
  \mathrel{#4}%
}
\makeatother
\DeclareMathOperator{\dep}{\not\! \perp\!\!\!\perp}
\newcommand\independent{\protect\mathpalette{\protect\independenT}{\perp}}
\def\independenT#1#2{\mathrel{\rlap{$#1#2$}\mkern2mu{#1#2}}}

\theoremstyle{definition}

\newcommand{\indep}{\rotatebox[origin=c]{90}{$\models$}}

\DeclareMathOperator*{\argmin}{arg\,min}

\usepackage[accepted]{icml2021}

\icmltitlerunning{Operationalizing Complex Causes: A Pragmatic View of Mediation}

\begin{document}
\twocolumn[
\icmltitle{Operationalizing Complex Causes: \\ A Pragmatic View of Mediation}




\begin{icmlauthorlist}
\icmlauthor{Limor Gultchin}{csox,turing}
\icmlauthor{David S. Watson}{stats}
\icmlauthor{Matt J. Kusner}{csucl}
\icmlauthor{Ricardo Silva}{stats,turing}

\end{icmlauthorlist}

\icmlaffiliation{csucl}{Department of Computer Science, University of College London, London, UK}
\icmlaffiliation{turing}{The Alan Turing Institute, London, UK}
\icmlaffiliation{stats}{Department of Statistical Science, University of College London, London, UK}
\icmlaffiliation{csox}{Department of Computer Science, University of Oxford, Oxford, UK}

\icmlcorrespondingauthor{Limor Gultchin}{limor.gultchin@gmail.com}

\icmlkeywords{Machine Learning, ICML}

\vskip 0.3in
]



\printAffiliationsAndNotice{} 

\begin{abstract}

We examine the problem of causal response estimation for complex objects (e.g., text, images, genomics). In this setting, classical \emph{atomic} interventions are often not available (e.g., changes to characters, pixels, DNA base-pairs). Instead, we only have access to indirect or \emph{crude} interventions (e.g., enrolling in a writing program, modifying a scene, applying a gene therapy). In this work, we formalize this problem and provide an initial solution. Given a collection of candidate mediators, we propose (a) a two-step method for predicting the causal responses of crude interventions; and (b) a testing procedure to identify mediators of crude interventions. We demonstrate, on a range of simulated and real-world-inspired examples, that our approach allows us to efficiently estimate the effect of crude interventions with limited data from new treatment regimes.
\end{abstract}

\section{Introduction}\label{sec:introduction}

Understanding causal mechanisms is a primary goal of scientific inquiry and a crucial prerequisite for planning effective interventions. However, the task of isolating and quantifying treatment effects is complicated by several obstacles. Fundamental questions of identifiability \citep{shpitser2008, correa2020calculus} and transportability \citep{Bareinboim2016} pose significant challenges to practitioners across a variety of disciplines. The problems are particularly acute in high-dimensional settings where interventions are rarely of the surgical or ``atomic'' sort envisioned by most authors in this area. For instance, genomic data contains rich information about the pharmacodynamic impact of drug therapies on disease activity. However, careful analysis is required to detect and operationalize these sparse signals, as causal effects are not defined in terms of direct interventions on, say, individual genes, but are instead propagated from a crude treatment (drug administration) on a complex object (the human transcriptome), which affects outcomes (disease activity) through several mediating pathways. Similar complexity arises in other fields, for instance when purported causes are social constructs like ``gross domestic product'' \citep{arnold:2020} or large-scale natural phenomena like ``El Ni\~{n}o'' \citep{chalupka2016unsupervised}.

\begin{figure}[t!]
    \centering
    \includegraphics[width=\columnwidth]{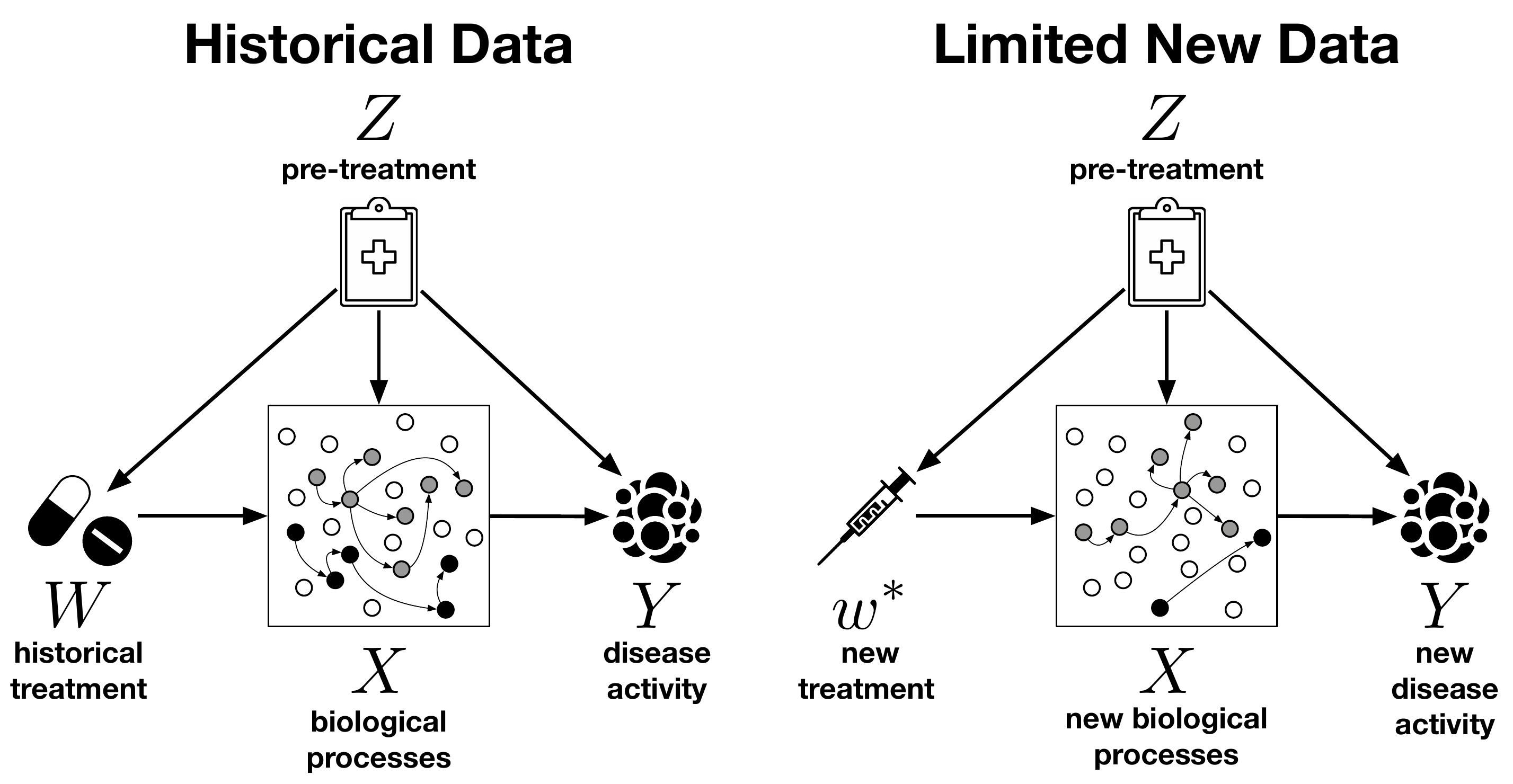}
    \vspace{-4ex}
    \caption{The complex cause problem setting. See text for details.}
    \vspace{-3ex}
    \label{fig:setup_intro}
\end{figure}

Despite a substantial and growing literature on causal inference (see Sect.~\ref{sec:problem}), existing theory largely fails to accommodate complex systems where the putative causes $X$ of an outcome of interest $Y$ have many internal components $(X_1, X_2, \dots, X_p)$ not amenable to perfect control. Using the notation of \citet{Pearl2000}, there is no clear, non-trivial, physical method for enacting $do(x)$, i.e. setting variable(s) $X$ to a particular value(s) $x$. Such cases are common in the natural and social sciences, to say nothing of text data and spatial processes captured at a coarse resolution. To continue with the medical example, researchers often design a therapy to target one or several hub genes in full awareness that this may spur unintended interactions with other biological processes. In such a study, researchers want to learn not just whether the drug is effective but how variability in patient response can be explained by elements of $X$ (perhaps combined with effect modifiers of pre-treatment variables $Z$). 

This gives operational meaning to the notion of $X$ as a \emph{cause} of $Y$: even if $do(x)$ is undefined, we are interested in framing the effect on $Y$ by some treatment $W$ where the design of $W$ comes from postulated or conjectured mechanisms triggered by $X$. Under the assumption that $X$ fully \emph{mediates} the actions $W$ (in a technical sense defined in Sect.~\ref{sec:problem}), an invariant relationship between $X$ and $Y$ under $W$ becomes a useful building block for predicting the outcomes of new interventions. Furthermore, understanding \emph{which} components of $X$ simultaneously covary with $W$ and $Y$ is of independent interest, as this may suggest new targeted interventions that operate on those elements of $X$.

In this work, we discuss a notion of \emph{pragmatic mediation} that provides a solution to the problem of prediction from new interventions. Our main contributions are threefold: (1) We formalize a general problem in which complex object $X$ causes outcomes $Y$ as a result of crude interventions $W$, with applications to domains with structured, high-dimensional data. (2) We describe an efficient method for estimating responses to new interventions, tractably marginalizing over the complex object $X$. (3) We propose a methodology for identifying practical causal mediation paths, which can provide insight into complex systems and suggest new hypotheses for future experiments.

\section{Problem Setup}\label{sec:problem}

Let $Y$ be an outcome of interest, and let $X$ be postulated \emph{causes} of $Y$, in the sense that hypothetical interventions on $X$ would alter the distribution of $Y$ \citep{Woodward2003a}. In the machine learning and artificial intelligence literature, this is typically operationalized in terms of the interventional distribution $p(y~|~do(x))$ \citep{Pearl2000}. In many domains, however, perfect control is ambiguous or unattainable \citep{vanderweele2013}. This is often the case when $X$ is a composition of more fundamental variables, as in the examples discussed in Sect.~\ref{sec:introduction}, as well in image and text data. We will explore the latter two in our experiments in Sect.~\ref{sec:experiments}.
\paragraph{Actionable variables and their use.}

In our setup, actions that change the distribution of $X$ are assumed to exist. We index them by $W$, allowing this to be a random vector. Pre-treatment variables $Z$, which are realized before $\{W, X, Y\}$, are also allowed. We say that $W$ are our \emph{actionable} variables, in the sense that in principle we can set them to exact values by an intervention.

By way of contrast, in the instrumental variable scenario, the target is $p(y~|~do(x))$, with $W$ acting as an instrument that is not fundamental to the estimand of interest. If $do(x)$ is not defined and we are primarily driven by policy questions (e.g., choosing an optimal value for $W$), then $W$ arguably makes the notion of $X$ as a cause redundant. For example, \citet{gelman:2009} suggests interpreting $X$ as little more than a qualifier for our actions $W$.
This is not satisfactory for the many applications where $W$ was chosen because we expect that changing $X$ will also change $Y$, even if this notion of  propagation is unclear when $do(x)$ is undefined. In particular, there are practical scenarios where assumptions about invariances involving $X$ aid the learning and prediction of policy outcomes. That is our motivation here.


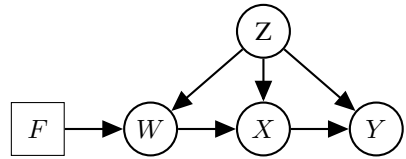
\begin{figure}

  \centering
  
  \begin{tikzpicture}
  \node[obs] (Z) at (3.0, 1.3) {Z};
  \node[plate] (F) at (0,0) {$F$};
  \node[obs] (W) at (1.5, 0) {$W$};
  \node[obs] (X) at (3.0, 0) {$X$};
  \node[obs] (Y) at (4.5, 0) {$Y$};
  \edge{F}{W}
  \edge{Z}{W};
  \edge{Z}{X};
  \edge{Z}{Y};
  \edge{W}{X};
  \edge{X}{Y};
  \end{tikzpicture}
 
 \caption{A DAG encoding independence assumptions in the set $\{F, W, X, Y, Z\}$, where random variables are circular vertices and intervention variable $F$ is a square vertex. This diagram captures conditional independencies assumed in our setup, but is not Pearlian, as $do(x)$ is undefined. We cover all members of the Markov equivalence class of this graph, including those with unmeasured confounding between $X$ and $Z$. We do not consider other unobserved confounders, though we discuss this in Sect.~\ref{sec:conclusion} as a direction for future work.
 \vspace{-3ex}}
 \label{fig:setup}
 
\end{figure}

\paragraph{Structural assumptions.}

Our goal is to predict $Y$ under intervention levels $w^\star$ of $W$ that we have not yet seen. Following \citet{dawid2020}, we introduce a \emph{regime indicator} $F$, which is not a random variable but instead indexes the conditional distributions $p_F(w~|~z)$, with values ranging over possible interventions on $W$. 
Following \citet{Pearl2000}, we use the symbol ``$ do(w)$'' $\in \mathcal F$ to mean the distribution in which $W = w$ given any $Z = z$. Our goal is then to predict $Y$ given $Z$ under $F = do(w)$, in particular reporting $\mathbb E_{F = do(w^\star)}[Y~|~z]$ for some new $w^\star$. In what follows, we will use the Pearlian notation $\mathbb E[Y~|~do(w), z]$ to represent the interventional distribution. Note that regime indicators can also accommodate non-atomic interventions -- e.g., idle or stochastic regimes \citep{correa2020calculus} -- although we will not make use of this in the sequel.

As is well-understood in the causal modeling literature, assumptions of independence between interventions and random variables can be represented by a directed acyclic graph (DAG). Our structural assumptions are encoded in Fig.~\ref{fig:setup}. This is \emph{not} a causal graph in the sense of \citet{Pearl2000}, as we are not making any claims about, say, lack of hidden common causes between $Z$ and $X$ or the applicability of $do$ interventions on all variables. Instead, the DAG represents conditional independence claims such as $F \indep ~Y~|~\{W, Z\}$, a statement interpretable as the lack of unmeasured confounding between $W$ and $Y$ given $Z$.\footnote{A similar device is used, for instance, in the proof of the back-door adjustment, Thm. 3.3.2 of \citet{Pearl2000}; and earlier graphical notions of unconfoundedness, e.g. Fig. 3.19 of \citet{Spirtes2000}.}

Of particular importance to what follows is the implication $\{F, W\} \indep Y~|~\{X, Z\}$. This conditional independence relationship, visually apparent from the $d$-separation in Fig.~\ref{fig:setup}, informs us that there is no direct effect of $W$ on $Y$. This invariance is the key point that pools together data collected at different values of $F$.
\vspace{-1.5ex}
\paragraph{Problem statement: learning with pragmatic mediation.}
\emph{Let $\mathcal F = \{f_1, f_2, \dots, f_m\}.$ Moreover, let $\mathcal D_l$ denote $m'$ ``labeled'' datasets 
\[\mathcal D_l \equiv \{(W, X, Y, Z)^{l_1}, \dots, (W, X, Y, Z)^{l_{m'}}\},\]
where $\mathcal L \equiv (l_1, \dots, l_{m'}) \subset [m]$, and let $\mathcal D_u$ denote $m''$ ``unlabeled'' datasets 
\[\mathcal D_u \equiv \{(W, X, Z)^{u_1}, \dots, (W, X, Z)^{u_{m''}}\},\]
where $\mathcal U \equiv (u_1, \dots, u_{m''}) \subset [m]$. In particular, $(W, X, Y, Z)^{l_i}$, $l_i \in \mathcal L$, denotes data collected under regime $f_{l_i}$, the analogous holding for $u_i \in \mathcal U$. Given $(\mathcal D_l, \mathcal D_u)$, the goal is to return an estimate of}
\begin{equation}
    f(w^\star, z) \equiv \mathbb E[Y~|~do(w^\star), z].
\label{eq:goal}
\end{equation}
\noindent The problem concerns evaluating outcomes under an intervention that sets $W$ to $w^\star$. The \emph{post-treatment} variables $X$ cannot be used as a basis for decision-making, but the invariances encoded by the lack of edges $\{F, W\} \rightarrow Y$ and $F \rightarrow X$ allow for predictions of policy outcomes even in the case where pairs $(w^\star, y)$ are not in our data. As is typical of causal inference problems, we require some assumptions regarding support of treatment values in the given data. In particular, we have the following:

\noindent   \textit{\textbf{Assumptions (identification and support)}. For all $z$ in the support of $p(z)$: (i) the distribution $p(x~|~w^\star, z)$ is identifiable from the distributions sampled by $\mathcal L \cup \mathcal U$; (ii) the support of $p(x~|~w^\star, z)$ is contained in the union of the support of $X$ in each dataset in $\mathcal L$.} 

Thus, in order to obtain $\mathbb E[Y~|~do(w^\star), z]$ from
\[
\int \mathbb E[Y~|x, z]p(x~|~w^\star, z)~dx,
\]
\noindent we must have some means of generalizing to $p(x~|~w^\star, z)$ from past data, including unlabeled data. Condition (ii) says we can learn the $\mathbb E[Y~|~x, z]$ factor across the support of $p(x~|~w^\star, z)$ using the datasets contained in $\mathcal L$. This assumption can be relaxed, provided we have some principled way to extrapolate beyond the regions of $(X, Z)$ covered by $\mathcal L$. 

In our experiments (see Sect.~\ref{sec:experiments}), we predict causal responses for new interventions with no observed outcomes but some (unlabeled) data on $X$. Such cases arise when, for instance, $Y$ takes a long time to be observed, or past interventions $w^\star$ took place targeting a different outcome variable. We will not constrain the functional relationship between $W$ and $X$ in any way, meaning that $p(x~|~w^i, z)$ need not contain any information about $p(x~|~w^j, z)$ for $i \neq j$.


This machinery operationalizes what we mean by $X$ being a cause of $Y$, even if we do not define $do(x)$. At the heart of causal inference is the notion of invariance under intervention, which can be exploited even if the putative causes of interest cannot be directly manipulated. We call $X$ a \emph{pragmatic mediator}, i.e. a set of variables that allows us to decompose a causal model for $W$ in $Y$ by a model (given $Z$) relating $W$ and $X$ only, and $X$ and $Y$ only, under a space of interventions $\mathcal F$. This bears little relation to counterfactual mediation \cite{vanderweele:2015} and demonstrates how restricted notions of mediation can be more useful than counterfactual ones in some contexts.



\paragraph{Related work.} Although invariance principles have long been cited in formal definitions of causality \cite{Spirtes2000, Pearl2000, Buhlmann2020}, they have recently found a new life in machine learning approaches that target more focused questions of practical interest. 

The Invariant Causal Prediction method of \citet{peters2016} -- later extended by \citet{Heinze2018} and \citet{gamella2020} -- exploits the assumption that $F$ does not directly cause outcome $Y$ except for (unknown) causal parents from a candidate pool of observable variables $X$. There the objective is to discover the causal parents as opposed to learning what happens when we marginalize over them. Likewise, Invariant Risk Minimization \citep{arjovsky2020invariant} exploits variability in $F$ to better learn the relationship between $X$ and $Y$ in a way that is robust to estimation errors due to spurious, unstable associations. Again, the main focus is on the use of $X$, here in a (non-causal) prediction problem.

Post-treatment variables are used to improve bandit optimization by factoring the arm space, where variables that are themselves targets of interventions exhibit some (at least partially known) structure \cite{lattimore:2016, sanghack:2018, dekroon:2020}. In contrast, we focus on the class of problems where there is little to be gained by exploiting the inner structure of $X$, as in many applied scenarios they are composite variables with ambiguous fine causal structure \cite{arnold:2020}.

Domain adaptation, in particular covariate shift, has a long tradition of being analyzed in the context of causal models (e.g., \citet{zhang:2013}). The emphasis of this literature is how to better cope with changes in distribution between, say, training and test regimes. Although techniques such as sample reweighting for improved statistical efficiency are relevant when dealing with multiple regimes, this will not be our focus here. Instead, in Sect.~\ref{sec:method}, we emphasize convenient parametrizations of our causal setup so as to facilitate marginalization over $X$ and parameter learning.

Finally, this work is particularly influenced by the literature on ambiguous or undefined interventions \cite{spirtes:2004, vanderweele2013, sanghack:2019} as well as causal abstractions and compositional data \cite{chalupka:2017, beckers2019abstracting, arnold:2020}. The idea of pragmatic mediators is essentially the grounding of a causal abstraction through imperfect interventions $W$ and the delimitation of its possible values as allowed by a ``causal dictionary'' $\mathcal F$. The ambiguity of $X$ requires explicit assumptions about the space of modifications that we expect to enact on $X$.

\section{Method}\label{sec:method}
Based on the assumptions introduced above, this section describes (a) an algorithm for learning expected outcomes $Y$ under crude interventions on $X$, operationalized as $F = do(w)$, conditioned on pre-treatment covariates $Z$; and (b) a procedure for interpreting the elements of $X$ that play a mediating role in the fitted model. 

\paragraph{Causal response estimation.}
We assume access to a set of features $\phi_i: \mathcal{X} \times \mathcal{Z} \rightarrow \mathbb{R}$. These features are \emph{candidate mediators}, moderated by covariates $Z$, which describe the outcome model for $Y$ as
\begin{align}
Y = \theta_0 + \sum_{i=1}^d \theta_i \phi_i(X, Z) + \epsilon, \label{eq:y}
\end{align}
where $\epsilon$ is an independent error term with $\mathds{E}[\epsilon] = 0$. 

Candidates may come from domain experts (e.g., experimentally validated regulatory pathways) or a data-driven approach (e.g., latent factors learned by an autoencoder). They represent a macro-level summary that clarifies what the existing $\mathcal F$ is able to modify in $X$ that simultaneously contributes to $Y$. For example, if $X$ describes a spatially-distributed object, like neural activations or environmental sensors, features $\phi_i$ can correspond to smoothing windows with localized information. If $X$ is a text document, $\phi_i$ may represent aggregate interpretable interactions of relevant entities, topics, and other parts of speech. The linear assumption is substantive but not especially restrictive, given a sufficiently flexible library of basis functions $\Phi$, which, as mentioned above, can be trained directly via neural networks or some other representation learning method.\footnote{Note that, though each $\phi_i$ is a function of $X$ and $Z$, we occasionally simplify notation by suppressing the dependence, writing $\phi_i$ for $\phi_i(X,Z)$ and $\Phi = \{\phi_i\}_{i=1}^d$.} 


Given Eq.~\eqref{eq:y}, it follows by the assumptions encoded in Fig.~\ref{fig:setup} and by linearity of expectation that
\begin{align}
    \mathbb{E}[Y \mid do(w), z] = \theta_0 + \sum_{i=1}^d \theta_i \mathbb{E}[\phi_i(X, Z) \mid w, z]. \label{eq:do}
\end{align}
We therefore propose a two-stage procedure to estimate $\mathbb{E}[Y \mid do(w), z]$: 
\vspace{-1ex}
\begin{enumerate}
    \item Learn $g_i(w,z) \equiv \mathbb{E}[\phi_i(X, Z) \mid w, z]$ for all $i$ via any black-box regression algorithm, and let $\hat{\textbf{g}}$ denote the $d$-dimensional vector of resulting expectations.
    \item Learn $\bm{\hat\theta} = \argmin_{\bm{\theta}} \mathbb{E}[(Y - \bm{\theta}^\top \hat{\textbf{g}})^2]$ via regularized regression \citep[e.g.  Lasso,][]{tibshirani1996regression}, to provide sparsity on $\bm{\theta}$ where supported by the data.
\end{enumerate} 
\vspace{-1ex}


The procedure is detailed in Alg.~\ref{alg:prediction}, where we consider the case in which labeled datasets are pooled together into a set with $n$ samples, and we learn a model for $p(x~|~w^\star, z)$ from unlabeled conditions with a single treatment level $w^\star$. This exploits the known structural relationship between $W$, $X$, $\Phi$ and $Y$. In particular, it represents the marginalization of $X$ directly in terms of $\E[\phi_i(X, Z)~|~w, z]$,\footnote{This can be even further simplified if we opt for product features of the shape $\phi_i(X, Z) \equiv \phi_{ix}(X)\phi_{iz}(Z)$, as in this case we have $\E[\phi_{ix}(X)\phi_{iz}(Z)~|~w, z] = \E[\phi_{ix}(X)~|~w, z]\phi_{iz}(z)$ \citep{kaddour2021}.} which avoids the density estimation problem of learning $p(x~|~w, z)$. 

There is a relation between this idea and methods for estimating non-linear causal effects in additive-error instrumental variable models based on (potentially infinite) basis expansions \cite{singh:2019,muandet:2020}. However, given the potential high-dimensionality of $X$ and the desire for interpretability, we favor dictionaries that are either hand-constructed or the result of adaptive algorithms. Moreover, although we have the option of fitting $\theta$ by regressing $Y$ directly on $\Phi$, we still favor the regression on $\hat{\textbf{g}}$ instead, as $\phi_i{x, z}$ is a random variable not observable at test time.

\begin{algorithm}[!t]
   \caption{Causal Response Prediction}
   \label{alg:prediction}
\begin{algorithmic}
 \Require Historic interventions $\{w_i, \Phi(x_i,z_i), z_i, y_i\}_{i=1}^{n}$, new intervention training set $\{w^\star, \Phi(x_j,z_j), z_j\}_{j=1}^{n^\star_{train}}$, 
 new intervention test set $\{{w^{\star}}', z_{j}'\}_{j=1}^{n^\star_{test}}$ \\
 \State \textbf{Historic Interventions}
 \State Learn $g_k(w, z) = \E[\phi_k (X, Z)~|~w, z]$ \\
 \Comment Stage 1, via any black-box model
 \State Learn $f(w,z) = \E[Y~|~\textbf{g}(w,z)]$ \\
 \Comment Stage 2, via an $L_1$-penalized model
 \\
 \State \textbf{New Intervention}
 \State Learn on train split 
 \State $g^\star_k(w^\star, z) = \E[\phi_k (X, Z)~|~w^\star, z]$ \\
 \Comment Stage 1, update for new intervention $w^\star$
 \State Predict on test split
 \State $\hat{y} = f(\textbf{g}^\star({w^\star}', z')) = \E[Y~|~do({w^{\star}}'), z']$\\
 \Comment Stage 2, predict using pre-learned $f$\\
 \Return $\hat{y}$
\end{algorithmic}
\end{algorithm}

\begin{algorithm}[!ht]
   \caption{Pragmatic Mediation Selection}
   \label{alg:mediation}
\begin{algorithmic}
 \Require Weights $\mathbf{\theta}$, training set $\{w_i, \Phi(x_i,z_i), z_i\}_{i=1}^{n}$, test set $\{w_i', \Phi'(x_i',z_i'), z_i'\}_{i=1}^{n'}$, one-sided paired difference test $c(\cdot)$, level $\alpha$, mediators $\mathcal{M} = \{\}$\\
 
\For {$\phi_{i} \in \Phi$}
    \If {$\theta_i \neq 0$} 
        \State Learn $g_i^0(z) = \E[\phi_i(X, Z)~|~z]$ on train split
        \State Learn $g_i^1(z, w) = \E[\phi_i(X, Z)~|~z, w]$ on train split
        \State Obtain residual $\epsilon_i^0 = \phi_i' - g_i^0(z')$ on test split
        \State Obtain residual $\epsilon_i^1 = \phi_i' - g_i^1(z', w')$ on test split
        \State $p = c(|\epsilon_i^0|, |\epsilon_i^1|)$
        \If {$p \leq \alpha$}
            \State Add mediator $\mathcal{M} = \mathcal{M} \cup \{\phi_i\}$
        \EndIf
    \EndIf
\EndFor  
\\
\Return $\mathcal{M}$
\end{algorithmic}
\end{algorithm}

\paragraph{Explainable pragmatic mediation.} Under the assumptions of our setup, we would like to provide practitioners with qualitative information on the estimated role of the candidate mediators. Informally, we say that $\phi_i(X, Z)$ is a \emph{causal pragmatic mediator} if and only if it covaries with $W$ and $Y$ simultaneously, with adjustments for $Z$ and the other candidate mediators depending on the scenario. More formally, causal pragmatic mediators satisfy two criteria: 
\vspace{-1ex}
\begin{itemize}
    \item[(i)] $\phi_i(X, Z) \dep W~|~Z$,
    \item[(ii)] $\phi_i(X, Z) \dep Y~|~\{\Phi_{\backslash i}, Z\}$
\end{itemize}
\vspace{-1ex}
\noindent where $\Phi_{\backslash i} \equiv \Phi \backslash \phi_i(X, Z)$. We will henceforth refer to (i) and (ii) as $\mathcal M$-criteria.

Another way of interpreting this is by saying that $W$ has a ``nonzero conditional total effect'' on $\phi_i$ for some $Z = z$ (that is, conditional association without adjusting for $\Phi_{\backslash i})$, while $\phi_i$ has a ``direct effect'' on $Y$ (conditional association, also conditioning on $\Phi_{\backslash i}$). 

This definition is entirely agnostic to any possible causal structure among the elements of $\Phi$, a structure which is itself indeterminate since $do(x)$ is not defined. Notice that the idea of combining a ``total'' effect ``into'' $\Phi$ with a ``direct'' effect ``out of'' $\Phi$ relates to settings where we may want to design new elements of $\mathcal F$ that ``short-circuit'' the mechanism, by directly targeting $\phi_i$ if this is at all possible and desirable in a particular domain.\footnote{For instance, ignoring $Z$ for simplicity: if there exists a ``Pearlian'' causal chain $X_1 \rightarrow X_2 \rightarrow X_3 \rightarrow Y$ in the system with no further edges, and we have a rich dictionary $\Phi$, (features of) neither $X_1$ nor $X_2$ alone would qualify as causal pragmatic mediators, while (features of) $X_3$ would, even if all interventions in $\mathcal F$ can only directly modify $X_1$ and $X_2$.}

Although the distinction is not crucial for prediction, causal mediators can provide valuable insights about what in $X$ characterizes the effect of $W$ on $Y$. For instance, if only a subset of regions of the brain respond to stimuli and predict some behavior, then novel interventions can be designed targeting just those regions with  detectable causal impact.

\begin{figure}[t!] \label{fig:decomp}
\centering
    \includegraphics[width=\columnwidth]{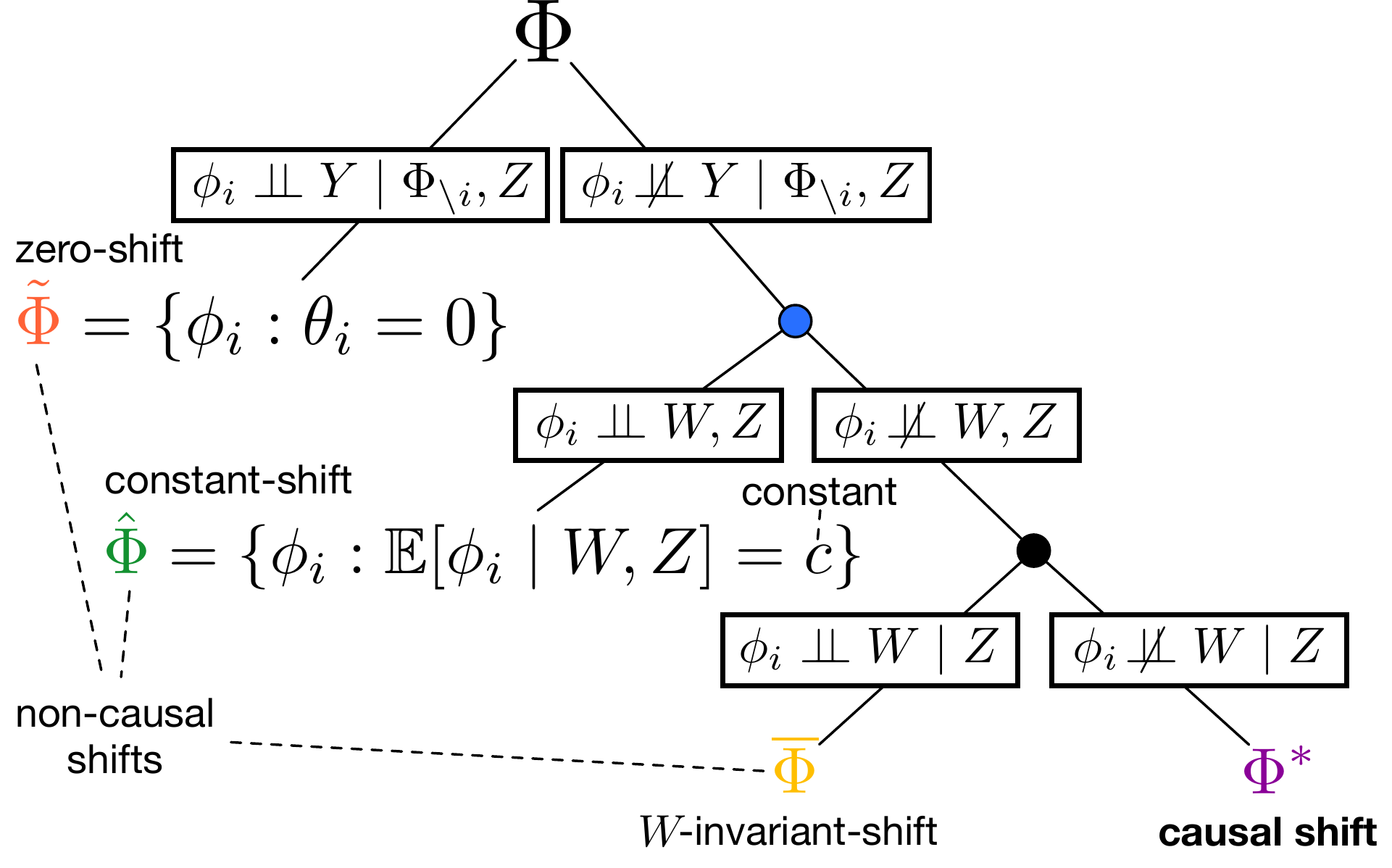}
    \caption{Recursive partition of $\Phi$ by how elements do or do not shift conditional average treatment effects. See text for details.}
    \vspace{-1ex}
    \label{fig:decomposition}
\end{figure}

The leaf nodes of the tree depicted in Fig.~\ref{fig:decomposition} correspond to candidate mediators with different functional roles. 
\vspace{-1ex}
\begin{enumerate}
    \item $\textcolor{CCoral}{\tilde{\Phi}} \equiv \{\phi_i: \phi_i \independent Y~|~\Phi_{\backslash i}, Z\}$. These candidates will receive zero weight in the linear formula described by Eq.~\eqref{eq:y} (and, hence, also Eq.~\eqref{eq:do}). That is, for each $\phi_i \in \textcolor{CCoral}{\tilde{\Phi}}, \theta_i = 0$.
    \item ${\textcolor{CGreen}{\hat{\Phi}}} \equiv \{\phi_i: \phi_i \not\in \textcolor{CCoral}{\tilde{\Phi}} \land \phi_i \independent W,Z\}$. In this subset, $\E[\phi_i~|~w, z]$ is constant for all $w$ and $z$. These terms will be absorbed into the intercept of the linear formula described by Eq.~\eqref{eq:do}. That is, $\mathbb{E}[Y \mid do(w), z] = \theta_0 + \sum_{\phi_i \in {\textcolor{CGreen}{\hat{\Phi}}}}\mathds{E}[\phi_i]~+~$``function of $w$ and~$z$''.
    \item ${\textcolor{CYellow}{\overline{\Phi}}} \equiv \{\phi_i: \phi_i \not\in \{\textcolor{CCoral}{\tilde{\Phi}} \cup {\textcolor{CGreen}{\hat{\Phi}}}\} \land \phi_i \independent W~|~Z\}$. These candidates will receive nonzero weight in Eq.~\eqref{eq:y}, but only through the $Z \rightarrow \phi_i \rightarrow Y$ path. They are invariant in $W$ and therefore, just like $\textcolor{CCoral}{\tilde{\Phi}}$ and ${\textcolor{CGreen}{\hat{\Phi}}}$, do not contribute to conditional average treatment effects $\mathds{E}[Y~|~do(w), z] - \mathds{E}[Y~|~do(w'), z]$. 
    \item $\textcolor{CViolet}{\Phi^*} \equiv \{\phi_i: \phi_i \in \Phi \backslash \{\textcolor{CCoral}{\tilde{\Phi}} \cup {\textcolor{CGreen}{\hat{\Phi}}} \cup {\textcolor{CYellow}{\overline{\Phi}}}\}\}$. Only this latter subclass satisfies the $\mathcal{M}$-criteria, picking out causal mediators $\phi_i$ on the $W \rightarrow \phi_i \rightarrow Y$ path. 
\end{enumerate}


This recursive partitioning of $\Phi$ immediately suggests a practical method for pragmatic mediation discovery. First, we perform our two-step estimation procedure. Then, for each $\phi_i$ such that $\theta_i \neq 0$, perform a conditional independence test against the null hypothesis $H_0: \phi_i \independent W~|~Z$.\footnote{In randomized trials, where $Z \independent W$ by design, this can be replaced by a marginal association test against $H_0: \phi_i \independent W$, for those $\phi_i$ which are non-trivial functions of $X$.} See Alg.~\ref{alg:mediation} for details.

There exists no uniformly valid conditional independence test for continuous conditioning variables \citep{Shah2018}. However, numerous nonparametric methods have been developed with good performance on real and synthetic datasets \citep{Heinze2018}. In our experiments, we use a simple nested regression procedure, in which we compare the absolute value of out-of-sample residuals for null and alternative models -- i.e., $g_{i}^{0}(z) = \E[\phi_i~|~z]$ and $g_{i}^{1}(z, w) = \E[\phi_i~|~z,w]$, respectively -- using a one-sided Wilcoxon rank-sum test.\footnote{Other tests could in principle be substituted here, e.g. the binomial test or $z$-test, depending on what assumptions one is willing to make about residual distributions. See \citep[Sect.~6]{Lei2018}.} If predictive accuracy significantly improves with the inclusion of $W$, then we reject $H_0$. Estimation and testing are performed on separate samples to ensure unbiased inference. The procedure can easily be modified to adjust for multiple testing.

\section{Experiments}\label{sec:experiments}
\begin{table*}[!t]
\vspace{-1ex}
\caption{General description of experimental setups.}
\vspace{1ex}
\resizebox{\linewidth}{!}{%
\begin{tabular}{@{}|l|l|l|l|@{}}
\toprule
 & \textbf{ImagePert} & \textbf{Humicroedit} & \textbf{DREAM5} \\ \midrule
\textbf{Z} & pre-perturbation image & original news headline (GloVe avg. vector) & baseline gene expression \\ \midrule
\textbf{W} & location of normal distribution for perturbation & new entity edit (GloVe vector) & transcription factor out-degree \\ \midrule
\textbf{X} & post-perturbation image & edited headline (GloVe avg. vector) & post-intervention gene expression \\ \midrule
\textbf{$\Phi$} & convolution windows over $X$ & funniness hypotheses  \citep{hossain-etal-2019-president} & change in kernel eigengene \\ \midrule
\textbf{Y} & intensity of pixels, linear combination of $\Phi$ & funniness score, via linear combination of $\Phi$ & linear combination of $\Phi$ \\ \bottomrule
\end{tabular}%
}
\label{tab:experiments_setup}
\end{table*}

In this section, we demonstrate our method in a variety of domains. We start with a simulated visual simulation task to provide a more concrete intuition for our approach. We then introduce a text data example where users are asked to edit news headlines to make them more humorous. Finally, we describe a genomics experiment where we simulate the effects of gene knockouts on the \emph{E. coli} transcriptome. The code to reproduce results can be found at \url{https://github.com/limorigu/ComplexCauses}.

\begin{figure}[!hbpt]
\vspace*{-1ex}
    \centering
    
    \subfloat[]{{\includegraphics[scale=0.3]{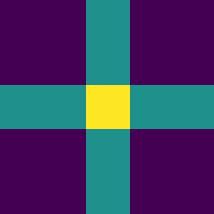}}}%
    \hspace{1ex}
    \subfloat[]{{\includegraphics[scale=0.3]{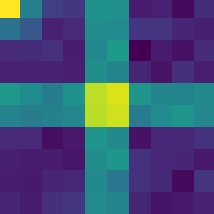} }}%
    \hspace{.4ex}
    \subfloat[]{{\includegraphics[scale=0.3]{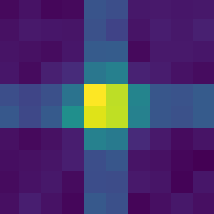} }}%
    \caption{Visual example for image perturbation dataset. (a) Example image ($z$). (b) Same image, post-perturbation ($x$, in response to $w$). (c) Same image under a new perturbation regime, which we leave for the test set ($x'$, in response to $w'$).}
    \label{fig:imgPert_simulation_setup}%
\end{figure}
\vspace{-3ex}
\subsection{Setup}
We have two primary goals: (a) causal response prediction, and (b) identification of causal pragmatic mediators. We describe the overall setup for all domains below.


\paragraph{Prediction.}

We assume access to $m - 1$ mutually independent historic training regimes with corresponding labeled datasets $\mathcal D_{l_1}, \dots, \mathcal D_{l_{m - 1}}$, where each $\mathcal{D}_{l_k} = \{(W_i, X_i, Y_i, Z_i)^{l_k}\}_{i=1}^{|\mathcal D_{l_k}|}$. Our goal is to learn $\E[Y~|~do(w^\star), z]$ for a new regime $F = do(w^\star)$ (e.g., a prospective intervention). In this regime, we are given access to limited labeled training data $\mathcal{D}_{l_{w^\star}} = \{(W^\star_i, X_i, Y_i, Z_i)^{l_{w^\star}}\}_{i=1}^{|\mathcal{D}_{l_{w^\star}}|}$ and more unlabeled training data $\mathcal{D}_{u_{w^\star}} = \{(W^\star_i, X_i, Z_i)^{u_{w^\star}}\}_{i=1}^{|\mathcal{D}_{u_{w^\star}}|}$, where $|\mathcal{D}_{u_{w^\star}}| \gg |\mathcal{D}_{l_{w^\star}}|$. This captures settings where measurements for $Y$ are expensive, delayed, or simply unrecorded. All methods are evaluated on a test dataset $\mathcal{T}_{w^\star} = \{(W^\star_i, Y_i, Z_i)^{t_{w^\star}}\}_{i=1}^{|\mathcal{T}_{w^\star}|}$.



Baseline methods that estimate $\E[Y~|~do(w^\star), z]$ can only make use of the labeled dataset $\mathcal{D}_{l_{w^\star}}$, as all regimes are mutually independent. However, by exploiting structural information $\Phi(X,Z)$, we are able to leverage the invariant $p(y~|~x, z)$ distribution from prior regimes. That is, we estimate $\textbf{g}$ and $\bm{\theta}$ from $\mathcal D_{l_1}, \dots, \mathcal D_{l_{m - 1}}$ and predict effects in new regimes using only $Z$ and $W$, so our method effectively treats $\mathcal{D}_{l_{w^\star}} \cup \mathcal{D}_{u_{w^\star}}$ as a single test set. 

We will compare our approach to multiple regression baselines that estimate $\mathbb{E}[Y~|~do(w^\star), z]$ as the proportion of labeled data for the new regime $\mathcal{L}_{w^\star}$ grows from $10\%$-$100\%$. Specifically, we consider models from four function classes: lasso regression (linear), support vector regression (SVR), random forest (RF), and gradient boosting (GB). Default hyperparameters are used throughout; see Appendix for details. We also note that other methods that seem to share similarity with our goal, such as co-training \citep{blum1998combining} and domain adaptation \citep{chen2011co}, would not be relevant baselines for comparison as they differ significantly from our work in two ways. (1) There is a two-stage functional decoupling arising from Eq.~\ref{eq:do} alongside variable decoupling that we aim to exploit by learning $g$ and then $f$; co-training does not involve such functional decoupling. (2) We can only leverage the first stage of the decoupling in a new domain. We are not aware of any domain adaptation method that accommodates this specific notion of adaptation.

As an additional check on our performance, we further consider 100 different settings of the target $Y$, by sampling 100 different parameters (i.e., weights $\bm{\theta}$) for its structural equations in all three tasks (for the image perturbation example we sample 1500 settings). By demonstrating consistent results across these trials, we illustrate that our method is robust to different configurations of the target variable.
\vspace{-1ex}
\paragraph{Explanation.} Our method is also able to find pragmatic mediators in the complex object $X$. By studying the high-level descriptions $\phi_i$ that (a) receive nonzero weight $\theta_i \neq 0$ in the sparse regression, and (b) reject $H_0: \phi_i \independent W|Z$ at some prespecified level $\alpha$, we can identify causal mediators of relevance. We report mediator discovery error rates for all experiments below. Significance levels for all tests were fixed at $\alpha = 0.01$, with $p$-values adjusted for multiple testing via \citet{Holm1979}'s method.



\subsection{Image Perturbation Simulation}
\label{sec:visual}

\textbf{Setup description.} 
Our first example is simulated and visual, which we hope will provide some intuition for the structure of this problem. We start with five possible pixel patterns ($Z$) and perform interventions by adding bivariate normal noise with location $W$. These treatments are influenced with some probability by $Z$. The resulting post-perturbation image ($X$) is then summarized via four different convolution windows, $\Phi(X, Z) = \{\phi_1,\phi_2,\phi_3,\phi_4\}$, where each $\phi_i$ corresponds to a quadrant of the image, and the convolution weights are indexed by the pattern corresponding to the original image $Z$. Finally, the intensity of the pixels over the whole image leads to an outcome ($Y$), given by a linear combination of $\Phi$. The generative model used to produce the simulation is described in Fig.~\ref{tab:visual_ex}.

\begin{figure}
\begin{equation}
\begin{aligned}
    t \sim&\; \text{Multinomial}(\bm{p})\\
    Z =&\; \mbox{pattern}_t \\
    W \sim&\; \text{Multinomial}(\Delta_t) \\
    f_w =&\; \begin{cases}
    \text{for } i=0 \text{ to } 1000: \\
        \hspace*{2em}\gamma \sim&\; \hspace*{-10em} \mathcal{N}(W,\textbf{I})\\
        \hspace*{2em} \text{if } (d_0,d_0) < \gamma < (d_n, d_n):\\
        \hspace*{4em}   f_w[\gamma] = f_w[\gamma] + \eta
    \end{cases}\\
    X =&\; Z+f_w+\mathcal{N}(0,0.5)\\
    \Phi =&\; \text{Convolution}_{t}(X) \\
    Y =&\; \boldsymbol{\theta}^\top \Phi + \mathcal{N}(0,0.1)
\end{aligned} \nonumber
\end{equation}
\vspace{-4ex}
\caption{Description of the generative model used in the experiment of Sect.~\ref{sec:visual}.
\vspace{-4ex}}
\label{tab:visual_ex}
\end{figure}

$\bm{p} = [0.2, 0.2, 0.2, 0.2, 0.2]$ defines the multinomial distribution from which we sample shape indicator $t$. $\Delta$ denotes a $5\times 4$ matrix, where each row is a simplex containing different probabilities for selecting $W$ values. $d_0=0 $ and $d_n=10$ define the dimensions of all images. The condition involving them and $\gamma$ checks whether the sampled location falls within the image size. $\eta$ is a perturbation parameter (fixed at 0.1 in our experiment) that is added to the sampled location if it passes the check above. This example is designed for demonstrative purposes, and $\phi_1$ is constructed to be the pragmatic mediator we intend to find, as it both varies with $W$ and has a nonzero coefficient ($\theta_1 = 0.7$) in the structural equation for $Y$. See full details in the Appendix. Fig.~\ref{fig:imgPert_simulation_setup} shows an example set of sampled images.

\begin{figure*}[!t]
    \centering
    \includegraphics[width=\textwidth]{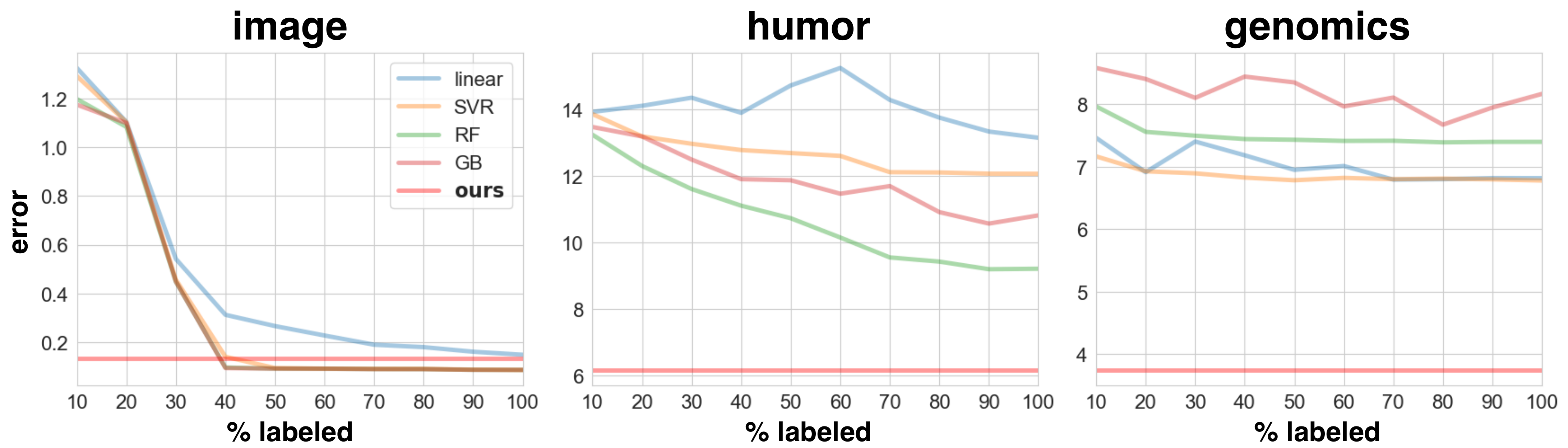}
    \vspace{-4ex}
    \caption{The mean squared error (MSE) between the estimated causal effect and the true causal effect  
    as a function of the amount of labeled data that is available in the new regime $do(w^\star)$.
    \vspace{-2ex}}
    \label{fig:MSE_results}
\end{figure*}

\begin{figure*}[!t]
    \centering
    \includegraphics[width=\textwidth]{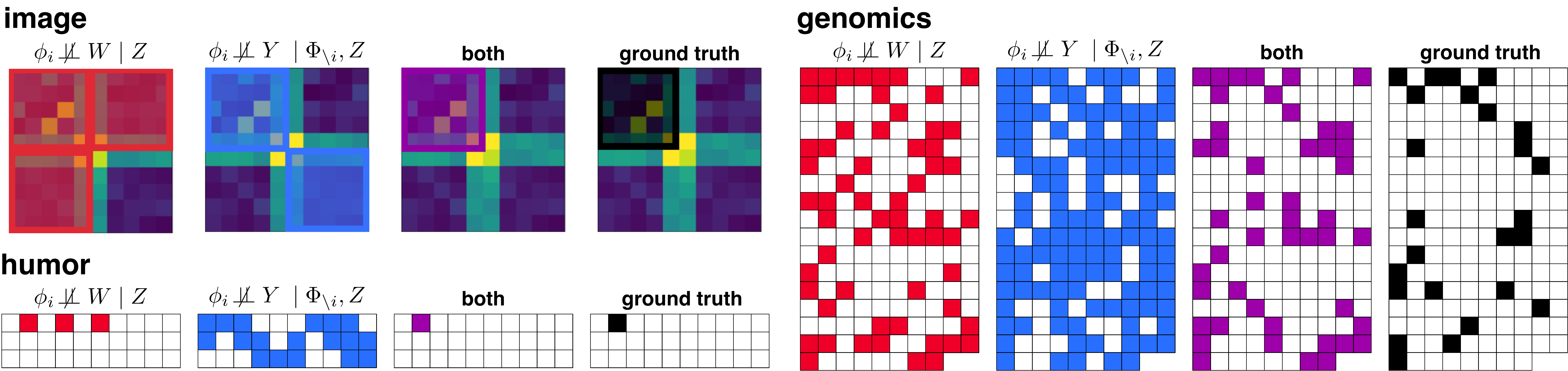}
    \caption{The $\phi_i$ selected by the mediation discovery method. Each set is identified as follows: \textcolor{CRose}{red} by a conditional independence test, \textcolor{CBlue}{blue} by sparse regression, and \textcolor{CViolet}{purple} those satisfying both tests (\textbf{black} are the true mediators). Note for the high-dimensional genomics dataset the \textcolor{CViolet}{$\phi_i$} are identified by only testing those \textcolor{CBlue}{$\phi_i$} selected by sparse regression to increase testing power.
    \vspace{-2ex}}
    \label{fig:found_phis}
\end{figure*}

\textbf{Results.}
The results of all methods on a new intervention $w^\star$ are presented in Fig. ~\ref{fig:MSE_results}. 
Additionally, Fig.~\ref{fig:found_phis} shows the process of mediator explanation for our method. The true mediator in this simulation, $\phi_1$, is indicated in black. 
Conditional independence tests identify three windows -- $\textcolor{CRose}{\Phi} = \{\phi_1, \phi_2, \phi_3\}$, indicated in \textcolor{CRose}{red} -- that vary with $W$ after conditioning on $Z$. We fit a lasso regression to estimate causal effects (see Eq.~\ref{eq:do}), selecting windows $\textcolor{CBlue}{\Phi} = \{\phi_1, \phi_4\}$, indicated in \textcolor{CBlue}{blue}. Finally, the intersection of these two sets, $\textcolor{CViolet}{\Phi^*} = \{\phi_1\}$, is our causal mediator, marked in \textcolor{CViolet}{purple}. Fig.~\ref{fig:MSE_results_diffY} presents performance over 1500 different samples of parameters in the structural equation of prediction target $Y$. It shows similar trends to the single $Y$ setting, where our method dominates performance by baselines until 30-40\% of labels are available. The mean squared error (MSE) in this simplified example is far smaller, and required more samples to make std. error scale accordingly. We further note that for the single $Y$ setting, we picked $\theta = \{0.7, 0, 0, -0.5\}$ to clearly  demonstrate the idea of pragmatic mediation. For Fig.~\ref{fig:MSE_results_diffY}, we instead sampled $\bm{\theta}$ from a distribution (See Appendix for details), which seemed to help the performance of some baselines, while adversely affecting others.


\subsection{Humorous Edits to News Headlines}

\textbf{Setup description.} As a second example, we consider a dataset from a computational humor experiment. Participants were given news headlines and asked to make single entity changes such that the resulting headline would be humorous  \citep{hossain-etal-2019-president}. This work was further extended into a SemEval2020 task, and full datasets were made publicly available.\footnote{See \url{https://www.cs.rochester.edu/u/nhossain/humicroedit.html}.} 

For our evaluation, we combine all listed datasets and define the following: original headline ($Z$), new word introduced by edit ($W$), revised headline ($X$). Following the analysis of \citet{hossain-etal-2019-president}, we carried out the following pre-processing procedures: 1. We generated clusters of edit words (granular $W$) by performing a $k$-means clustering on GloVe vector representations \citep{pennington2014glove} of each edit word, with $k=20$. The aim was to reduce the space of possible interventions to topics rather than individual words, for the purpose of defining data subsections as historic and new intervention splits. We used the resulting cluster label to create these splits. 2. We created 30 high-level descriptions $\phi$ for this setting (full description in the Appendix). One can think of $\Phi$ in this scenario as hypotheses to explain the funniness of an edited headline ($X$). 3. Computational humor is known to be a difficult domain for direct prediction tasks. For the illustrative purpose of this paper, we generated funniness scores for the outcome variable $Y$ as a linear combination of $\Phi$ with additive noise. A random third of the coefficients are assigned a value of 0, with the rest sampled from a uniform distribution $\mathcal{U}(-1,1)$.

\begin{figure*}[!t]
    \centering
    \includegraphics[width=\textwidth]{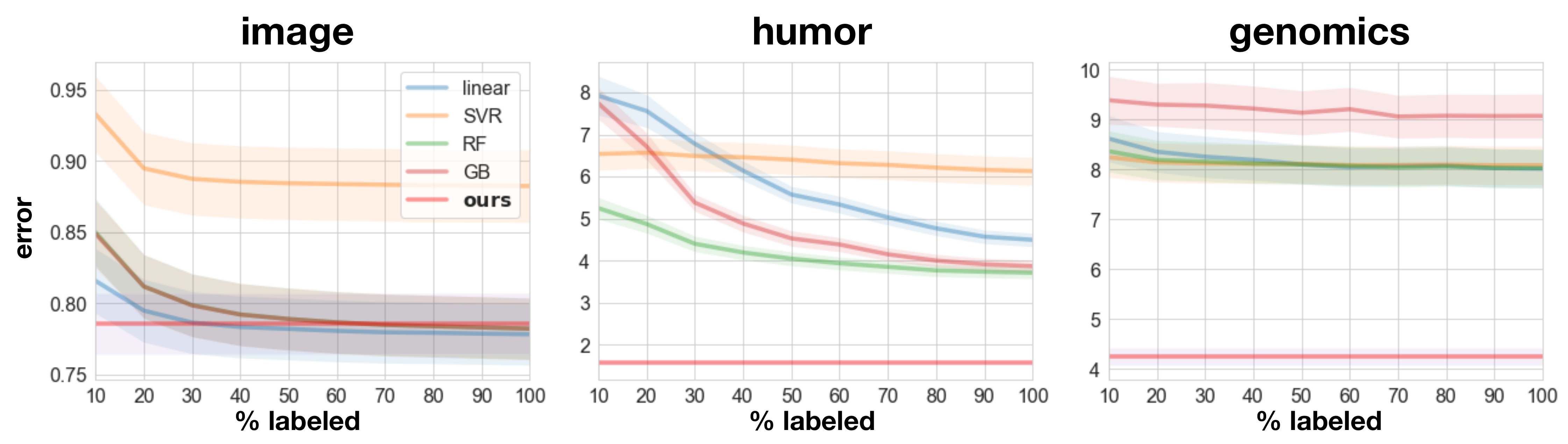}
    \vspace{-4ex}
    \caption{The mean squared error (MSE) between the estimated causal effect and the true causal effect  
    as a function of the amount of labeled data that is available in the new regime $do(w^\star)$. Means and std. error are over 1500 for the Image Perturbation experiments, and 100 for the other two, different configurations of $\bm{\theta}$, the parameters in the structural equations giving rise to $Y$.
    \vspace{-3ex}}
    \label{fig:MSE_results_diffY}
\end{figure*}

\textbf{Results.}
The results of our estimation method of the outcome $Y$ for a random new intervention $w^\star$ are presented in Fig.~\ref{fig:MSE_results}. As can be seen, we achieve a MSE of $5.33$, well below alternative estimation methods of $\E[Y~|~do(w^\star),z]$. Furthermore, our method correctly identified the mediator $\phi_2$ in this setting, see Fig.~\ref{fig:found_phis}. Fig.~\ref{fig:MSE_results_diffY} provides another angle on the quality of predictions with our method by examining results over 100 trials with different configurations of $\bm{\theta}$. We can clearly see that our method still outperforms the baseline alternatives, and sees little variation in performance across parameter values, as can be seen by the small std. error bars.

\subsection{Gene Knockouts}

\textbf{Setup description.} As a final experiment, we consider semi-simulated gene knockouts based on data from the Dialogue for Reverse Engineering Assessments and Methods (DREAM) challenge \citep{Marbach2012}. The \emph{E. coli} transcriptome published as part of the DREAM5 challenge comprises a $805 \times 4511$ gene expression matrix, with 334 candidate transcription factors.\footnote{See \url{http://dreamchallenges.org/project/dream-5-network-inference-challenge/}.} We use GENIE3 \citep{Huynh-Thu2010}, a leading gene regulatory network inference algorithm based on random forests, to fit the 4177 structural equations that govern this system. We treat the resulting model as our ground truth SCM. 

We simulate $n = 10^4$ samples of baseline expression data for the transcription factors from a multivariate Gaussian distribution with parameters estimated via maximum likelihood. These values are then propagated by GENIE3 to downstream variables, resulting in a complete set of baseline expression data ($Z$). We simulate 10 gene knockout experiments, summarized by the out-degree of the corresponding transcription factor ($W$). Post-intervention expression is once again simulated by GENIE3 ($X$). We treat each subnetwork of at least 10 genes as a pathway, and summarize its expression by taking the first kernel principal component \citep{kpca} of the corresponding submatrix, i.e. the kernel eigengene. The difference between post- and pre-intervention eigengene expression for all 168 modules that meet this dimensionality criterion constitutes our high-level summary ($\Phi$). Modules are subsequently ranked by their Spearman correlation with $W$. The top and bottom 25 are assigned nonzero weight in a linear simulation of outcomes $Y$, with standard normal noise. More details can be found in the Appendix.

\textbf{Results.} Results for a random new gene knockout are presented in Fig.~\ref{fig:MSE_results}. The sparsity of this problem poses a particular challenge for baseline regression methods, which could potentially be mitigated with further tuning. In addition to achieving low MSE on the test set, our method additionally recovers 92\% of all true mediators, with an overall accuracy rate of 85\%. Most of the errors in this example appear to derive from false positives in the lasso regression, which could likely be improved with more cautious tuning of the $\lambda$ parameter that controls model sparsity. As can be seen in Fig.~\ref{fig:MSE_results_diffY}, the same trends remain in place when repeating the experiment over 100 different configurations of $\bm{\theta}$.


\section{Conclusion}\label{sec:conclusion}
In this work, we proposed a general problem setup with applications in various fields of scientific study and policy design. We showed its relevance to different modalities and domain subjects, and developed a general estimation framework. This enables the study of causal effects of crude interventions on high-dimensional, complex objects that impact an outcome of interest via some high-level mediator(s). Our approach is useful when one wishes to estimate the effects of new interventions for which little labeled data is available. We further showed how such a method can illuminate the underlying causal structure governing the process by identifying pragmatic mediation pathways between the complex objects and the outcome. Future work could extend this approach to various tasks, including: estimation of causal effects in response to high-dimensional and/or soft interventions; hypothesis design via search for interventions predicted to achieve the outcome of interest; and prediction and mediation analysis with latent variables or partial knowledge of the true causal graph. 
\paragraph{Limitations.} 
We see this work as a first step in the study of complex causes and crude interventions, which are distinct from the atomic, soft, or stochastic interventions that have been previously studied. Though our method is focused on a particular problem setup, we have argued that a wide variety of problems share a similar structure. Additional work could make the method more applicable in settings with unobserved confounders, or where known relationships within objects (e.g., spatial characteristics) could be exploited for greater sensitivity. Other interesting extensions of this work could examine cases where $X$ does not fully mediate $W$, or when the set of abstract features $\Phi$ is not fully known. For the former, we believe that if the direct effect of $W$ on $Y$ is sufficiently weak, it should still be possible to exploit $X$ as mediator. For the latter, we envision adding smoothness constraints or parametric assumptions on $\Phi$, or simultaneously learning the abstract features, as in \citep{xu2020learning}.

\section*{Acknowledgements}
This work was partially supported by ONR grant 62909-19-1-2096 to DSW and RS.


\bibliography{bibliography}
\bibliographystyle{icml2021}
\onecolumn
\icmltitle{Supplementary Materials}
\section{ADDITIONAL EXPERIMENTAL DETAILS}
\raisebox{-1pt}{We provide additional details on the conducted experiments below.}
\subsection{Method, evaluation task and baselines}
\begin{figure}[!hbpt]
    \centering
    \includegraphics[scale=0.4]{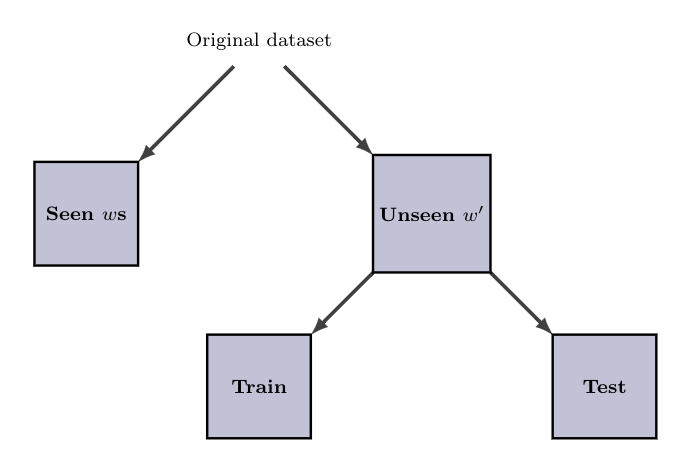}
    \caption{Description of dataset splits used in different parts of our experiment, corresponding to items 1-3 in \ref{item:experiment_desc}.}
    \label{fig:Dataset_desc}
\end{figure}

We tested our estimation method with three different setups: an image-based simulation, an experimental text dataset, and a naturally-simulated experimental genomics dataset. We considered the same evaluation task for all setups:
\begin{enumerate}
    \item Using a train set, with various ``seen'' $W$ values, we train a model for phi prediction, and use $\E[\Phi~|~W,Z]$ to fit a lasso regression to predict Y, giving rise to $\E[Y~|~\hat{\Phi}]$ predictions.
    \item Next, we see a train split from a test set, corresponding to an ``unseen'' intervention $w'$, on which we relearn $\E[\Phi~|~w',Z]$
    \item Finally, we test Y predictions on a test set of the unseen $w'$ test set, using our relearned $\E[\Phi~|~w',Z]$ model, and our previously trained $\E[Y~|~\hat{\Phi}]$ Lasso regression models. Thus, for a test split of the test set, we predict for Y for previously unseen ${w',z}$ pairs, i.e. $\E[Y~|~do(w'), Z']$.
\end{enumerate}\label{item:experiment_desc}
 For further clarity, we provide a visual description of datasets used in different stages of the method above in Figure \ref{fig:Dataset_desc}. We compared our estimation accuracy, via Mean Squared Error, as Y labels become available in the unseen $w'$ regime (10-100\% of labels). We record our loss against four baselines estimating the same quantity: 1. linear model, as a Lasso regression with cross validation to pick the coefficient $\lambda$ on the regularization term in the range [0.05,1], 2. SVM predictor with default parameters from \citep{scikit-learn}, 3. Random Forest regression with default parameters from \citep{scikit-learn}, aside for specifying 5 minimal samples in split, and 4. Gradient Boosting with default parameters from \citep{scikit-learn}, aside for maximum depth which was set to 5. We provide code to reproduce our results alongside this document. We also tested a Multi Layered Perceptron baseline, but found it to perform much worse than alternatives without dedicated tuning, and subsequently did not include it in the results.

For the prediction task we include two settings: one for a single configuration of the target $Y$, and one where we present average performance over 100 different settings of the parameters $\theta$ in the structural equation giving rise to $Y$. Since our test set on which we report result is rather small, for the first setting we average results of the baselines over 10 different shuffles of the dataset on which we train and report result. We do this for the baselines and not for our own method as the baselines have access to 10 different proportions of the target Y, which involved small number of samples and largely varying performance based on the ordering of the dataset. Our method makes no use of labels, and thus is not vulnerable to this variability in performance. However, when we conduct the experiment over 100 different $Y$ configurations, the existence of ample different examples of data better accounts for this variability in performance based on sample ordering. Thus, we simply report results for a single data ordering, and report the mean and standard error over 100 different $Y$ parameters settings instead.

\subsection{Dataset construction and models' training}
\subsubsection{Image Perturbation}
The image perturbation is a simulation dataset that was put together to demonstrate the key ideas of the work. We create a dataset of 10,000 examples, images of size 10X10, equally made up of 5 pixel patterns: cross, square, crossing diagonal, pyramid and diamond. Patterns were allocated for each index by sampling one of 5 pattern indicators from a multinomial with equal probability for each shape. Next, the pattern indicator $Z$ also served to select one of 5 set of probabilities that were used to seed a multinomial from which a perturbation pattern $W$ was picked.

The perturbation was put together via a procedure that used the indicator W as a location to be used in a Multivariate Normal distribution with a two dimensional identity covariance matrix, $I(2)$. For each example in the dataset, we sampled 1000 tuples (x,y) from the Multivariate Normal described, checked whether they fell within the image size (i.e. $\geq (0,0)$ and $<(10,10)$). Each time such tuple fell within the borders of the image, a perturbation of size 0.1 was added to the location (x,y) in the image.

This perturbation regime was added as a mask to the original image reflected by $Z$, to create the post-perturbation $X$. For the construction of $\phi$s, we created five different 1-d convolution transformations, with randomly initialized weights, which will be indexed by $Z$ and applied to $X$. There will be 4 resulting $\phi$s for each image, each corresponding to the 4 quadrants of the image. $\phi_1$ most clearly varies with $W$, with potential little effect on $\phi_2$ and $\phi_3$, but $\phi_4$ should see close to no effect in response to $W$, based on perturbation locations. Finally, Y was constructed as a linear combination of $\phi$s, with the weights [0.7, 0, 0, -0.5] applied to them, and added Gaussian noise $N(0, 0.1)$.

Finally, for the test set with an unseen perturbation pattern $w'$, 2,000 examples that were featured in the training set of size 10,000, were used with a different perturbation, with location $w'=5$. Previous perturbations ranged from 0-3. For specific structural equations corresponding to this description, see Figure 5 in the manuscript.

For the $g$ model, estimating $\E[\Phi~|~w',Z]$, we train a Multi-task MLP with 3 hidden layers, 512 hidden dimensions, and the ReLU activation function. We use 100 epochs and 50 epochs to train the first-stage model $g$ for seen w, and for the unseen w' train split respectively. We use the Adam optimizer, with learning rates of 0.002 and 0.001 used for optimization of each stage respectively. We use a train batch size of 400, and test batch size of 100. We set the seed at 42. We also include the code to reproduce the results, see code files for any additional hyperparameter setting.

Finally, for the robustness to settings of $\theta$ experiment (Figure 8), we had to sample the weights from a distribution such that we can repeat the process 1500 times. We chose a uniform distribution $\theta \sim Unif(-0.3, 0.3)$, and added noise to $Y$ from a $N(0, 0.01)$. We chose those such that similar stats of $Y$ can be achieved, compared to the single $Y$ setting experiment. See generation code in $\text{Python\_Img\_Humor/data\_generation\_notebooks/ImgPertSim.ipynb}$.

\subsubsection{Humor Micro Edits}
The construction of the humor micro edits dataset closely followed the analysis and description in \cite{hossain-etal-2019-president}. The original datasets provided there already contained original headlines (which we used as $Z$ for our purposes), edit words ($W$ for us) and humorous post-edit headlines ($X$)\footnote{Access to original dataset at \url{https://www.cs.rochester.edu/u/nhossain/humicroedit.html .}}. Following the analysis in the paper, we chose to represent each of these sentences and words using their pre-trained 6 Billion-token GloVe word-embedding vector representations, trained originally on 2014 English Wikipedia and Gigaword 5\footnote{Available for download at \url{https://nlp.stanford.edu/projects/glove/}.} \citep{pennington2014glove}. We only included examples in which all words were correctly identified in the pre-trained word embedding, following a standard cleaning procedure (see code for exact details).

$Z$, $W$ and $X$ where does based on vector representation of the original dataset. There are three additional steps we have taken to compose the final dataset. First, we clustered the edit words using K-means clustering with K=20 (implemented via Scikit-learn), following the same procedure carried out in \cite{hossain-etal-2019-president}. We used the labels of these clusters to create the training and test split, such that the test set included edit words from one cluster we left out to function as an unseen intervention $W'$. The unseen intervention was chosen to be one of the larger 5 clusters, to ensure enough training examples for estimation exist. The cluster that was randomly chosen for the results shown in the paper is cluster 11.

Next, we constructed high-level descriptions of the intervention and its implications on $X$ as $\phi$. $\Phi = \{\phi\}_{i=1}^{30}$, and each one was inspired by analysis and hypotheses from \cite{hossain-etal-2019-president}: 
\begin{enumerate}
    \item $\phi_1$: Length of resulting edited sentence \textit{(does not vary with $w$)}
    \item $\phi_2$: Mean cosine distance between GloVe vector of edit word and the rest of words in sentence \textit{(varies with $w$)}
    \item $\phi_3$: Location index of replaced word \textit{(does not vary with $w$)}
    \item $\phi_4$: Sentiment polarity of edit word, using the pre-trained sentiment processor from \citep{qi2020stanza}\footnote{Full usage details available at \url{https://stanfordnlp.github.io/stanza/sentiment.html .}} \textit{(varies with $w$)}
    \item $\phi_5$: Sentiment polarity of resulting sentence, using the pre-trained sentiment processor from \citep{qi2020stanza} (does not vary with $w$)
    \item $\phi_6$: Cosine distance between GloVe vector of edited word and GloVe vector of original word \textit{(varies with $w$)}
    \item $\phi_7$-$\phi_{10}$: Cosine distance of GloVe vector of edit word from neighboring words (2 preceding, 2 succeeding) \textit{(does not vary with $w$)}
    \item $\phi_{11}$-$\phi_{30}$: Distance of mean GloVe embedding of final sentence from clusters' centroids \textit{(does not vary with $w$)}
\end{enumerate}

The set of $\phi$s, which all correspond to different data types, were all scaled to have 0 mean and unit variance, to make them more comparable. Finally, following $\Phi$ as defined above we constructed an outcome variable $Y$ as a linear combination of $\Phi$, with added noise sampled from $N(0, .5)$. We sampled weights for each $\phi, \theta \sim U(-1,1)$, while keeping a random third of the weights at 0. Additionally, we ensured at least one of the $\phi$s varying with $W$ is also zeroed out, to have diversity of all possible cases present.

For the $g$ model, estimating $\E[\Phi~|~w',Z]$, we train a Multi-task MLP with 3 hidden layers, 512 hidden dimensions, and the ReLU activation function. We use 700 epochs and 100 epochs to train the first-stage model $g$ for seen w, and for the unseen w' train split respectively. We use the Adam optimizer, with learning rates of 0.002 and 0.001 used for optimization of each stage respectively. We use a train batch size of 400, and test batch size of 100. We set the seed at 42. We also include the code to reproduce the results, see code files for any additional hyperparameter setting.

\subsubsection{Gene Knockouts}

Our GENIE3 model follows the instructions of \citet{Huynh-Thu2010}, who scale all genes prior to analysis and fit a series of random forest regressions predicting the expression of each "downstream" gene as a function of the 334 candidate transcription factors (TFs). Each forest contains 1000 trees, with $mtry = \sqrt{334}$. The adjacency matrix is computed using the impurity importance measure originally proposed by \citet{Breiman2001}. 

We simulate TF data from a multivariate Gaussian distribution with parameters estimated via maximum likelihood. This matrix is then propagated through our GENIE3 model to simulate expression values for downstream genes, with random Gaussian noise $\mathcal{N}(0, \sigma^2)$, where $\sigma$ is the RMSE of the corresponding random forest on out-of-bag data. This data -- TFs and downstream genes -- together comprise the matrix $Z$ of simulated baseline \emph{E. Coli} gene expression.

We sort TFs by outdegree and simulate a knockout experiment on the top ten by replacing their values with a scalar 1 unit less than the observed minimum for each (how much less is irrelevant, as random forests are invariant to monotone transformations). We record the outdegreee of these TFs as $W$ and the resulting expression matrix as $X$. 

To compute $\Phi$, we filter out all TFs with outdegree less than 100 and treat each of the remaining 168 TFs as the hub of a module. For downstream genes, module membership is determined by whether the given TF was assigned importance of at least 10 in the GENIE3 adjacency matrix. For each module, we compute the first kernel principal component on a subsample of $n = 1000$ using a radial basis function with default bandwidth given by the median Euclidean distance. These weights are then used to project the remaining data $Z$ and $X$ into the latent space. We define $\Phi$ as the difference between pre- and post-intervention expression values for the kernel eigengene. We proceed to estimate a series of $\E[\phi_j~|~Z, W]$ regressions on a training set comprising 8 of 10 $w$ values using random forests with 500 trees and $mtry = p/3$, where $p$ is the number of genes in a given module.

Each $\phi_j$ is sorted by its association with $W$ using Spearman correlation. $Y$ is then simulated as a linear function of the top and bottom 25 $\phi$'s, with nonzero weights drawn from $\mathcal{N}(\pm{4}, 1)$, where $\pm{4}$ denotes that the amplitude is multiplied by $-1$ with probability 0.5. A lasso regression $\E[Y~|~\hat{\Phi}]$ is fit to the aforementioned training set, with $L_1$ penalty $\lambda$ selected via 10-fold cross-validation. Since, by construction $Z \independent W$ in this experiment, the conditional independence tests can be replaced by a marginal independence test. We use the Spearman correlation to measure the association between $W$ and each $\phi_j$ using the training set.

\end{document}